\documentclass{article}

\usepackage{microtype}
\usepackage{graphicx}
\usepackage{booktabs}
\usepackage{amsmath}
\usepackage{amssymb}
\usepackage{mathtools}
\usepackage{amsthm}

\usepackage{hyperref}

\usepackage[preprint]{icml2026}

\usepackage[utf8]{inputenc}
\usepackage[T1]{fontenc}
\usepackage{url}
\usepackage{amsfonts}
\usepackage{nicefrac}
\usepackage{xcolor}

\usepackage[capitalize,noabbrev]{cleveref}

\icmltitlerunning{SAFAARI: Schema-Aware Framework for Accelerated Advertiser Response Intelligence}

\begin{document}

\twocolumn[
\icmltitle{SAFAARI: Schema-Aware Framework for\texorpdfstring{\\}{: }Accelerated Advertiser Response Intelligence}

\begin{icmlauthorlist}
\icmlauthor{Bhanu Teja Rangaraju}{amzn}
\icmlauthor{Chandan Kumar}{amzn}
\end{icmlauthorlist}

\icmlaffiliation{amzn}{Amazon, Seattle, USA}

\icmlcorrespondingauthor{Bhanu Teja Rangaraju}{ibhanu@amazon.com}
\icmlcorrespondingauthor{Chandan Kumar}{chndank@amazon.com}

\icmlkeywords{Agentic AI, Multi-Agent framework, LLM, Text-to-SQL, Schema Linking, Custom Evaluation, Evaluation Metrics, Virtual Assistant, Customer Support}

\vskip 0.3in
]

\printAffiliationsAndNotice{}

\begin{abstract}
The evolution of customer support systems is rapidly advancing with agentic chatbots, yet these systems face significant limitations when accessing enterprise data without predefined API endpoints. This paper presents SAFAARI (Schema-Aware Framework for Accelerated Advertiser Response Intelligence), a multi-agent framework that addresses the critical bottleneck of schema linking in Natural Language to SQL (NL-to-SQL) systems through specialized content, metadata, and orchestration agents. We also introduce SEAL (Schema Evaluation and Accuracy in Language\allowbreak-to-SQL), a novel composite metric that holistically evaluates system performance while penalizing inconsistent results. Through systematic experimentation with five feature set configurations, SAFAARI achieves an 81.66\% SEAL score (6.65\% improvement over baseline), with notable gains in datapoint accuracy (5.51\%) and schema-linking precision (4.69\%). The framework's effectiveness is validated through human-in-the-loop evaluation with domain experts, which proves its adaptability across diverse support domains. By automating the labor-intensive process of schema linking and query generation, our framework demonstrates 8× reduction in development time while maintaining high accuracy. The solution streamlines API development and enhances self-service capabilities, particularly benefiting customer support enterprises with complex data ecosystems.
\end{abstract}

\section{Introduction}
Agentic chatbots rely on three core components: knowledge bases, orchestration layers, and function calls. However, maintaining APIs for tool calls presents significant challenges, especially in diverse domains like advertising. NL-to-SQL systems offer a solution by enabling dynamic database access without predetermined API endpoints. This approach comprises of three key processes: datapoint identification, schema linking, and SQL generation. Schema linking, which connects natural language to database structures, presents a critical challenge due to the gap between human queries and database terminology. Our research introduces advanced schema linking techniques that resolve fundamental limitations in current systems.

Recent studies have identified several critical areas requiring improvement in schema linking. Error analysis conducted on prominent benchmarks like SPIDER and BIRD datasets reveals that schema-linking errors account for 29\%-49\% of all errors, significantly higher than other categories such as join operations (21\%-26\%) and nested queries (less than 20\%)~\cite{shi2024survey}. This statistical evidence underscores the urgent need for innovation in schema linking methodologies. This paper presents our contributions to advancing schema linking in text-to-SQL systems, with particular emphasis on addressing the challenges of complex schema and insufficient benchmarks. Our work aims to bridge the gap between academic research and practical applications, providing more robust and efficient solutions for implementing NL-to-SQL capabilities in agentic chat-bots. By focusing on schema linking innovations, we seek to enhance the overall performance and reliability of text-to-SQL systems, ultimately improving the capability of agentic chat-bots to handle complex database queries in real-world applications.

\section{Related Work}
The evolution of text-to-SQL systems can be categorized into three major stages. Initially, rule-based and template-based methods dominated the field~\cite{zelle1996learning}. Subsequently, deep learning approaches emerged, with sequence-to-sequence models becoming the mainstream solution, automatically mapping natural language inputs to SQL outputs without requiring intermediate steps like semantic parsing~\cite{shi2024survey}. In the current era, large language models (LLMs) have dramatically transformed the field, showcasing remarkable capabilities in understanding complex natural language questions and generating accurate SQL statements~\cite{zhang2024benchmarking}. Despite these advancements, significant challenges remain in handling complex cross-domain queries, particularly with real-world database schema~\cite{singh2025survey}.

Past work in the area shows that schema linking is a foundational component in text-to-SQL systems, responsible for identifying the relevant portions of a database schema needed to answer a natural language query~\cite{glass2025extractive}. Researchers have established that placing the entire schema in the prompt for an LLM can be computationally prohibitive, especially for models with smaller token windows, and even when technically possible, it remains expensive and potentially detrimental to model performance~\cite{glass2025extractive}. A novel approach is introduced to adapt decoder-only LLMs for schema linking that is both more computationally efficient and more accurate than generative approaches. The critical role of schema linking is emphasized by noting that ``the accuracy of initial schema linking significantly impacts subsequent SQL generation performance''~\cite{yuan2025knapsack}. Furthermore, researchers have highlighted the significant performance gap between text-to-SQL models on benchmarks versus real-world databases, particularly for natural language questions requiring complex filters and joins~\cite{nascimento2025text}. The work proposes database keyword search (KwS), as a strategy, to improve schema linking precision and recall, demonstrating how the combination of a KwS platform's data dictionary with a dynamic few-shot examples strategy can significantly enhance the schema linking process. The real-world applicability of NL-to-SQL systems is further emphasized to illustrate how natural language understanding can be applied to complex customer service routing problems~\cite{pi2024contact}. The work emphasizes the value of sophisticated language processing systems in practical business contexts, where understanding the complexity of natural language inputs is crucial for appropriate task routing and successful resolution.

In parallel, there has been significant research in exploring different agentic architectures to solve the schema linking problem. MAC-SQL, a novel LLM-based multi-agent collaborative framework was designed to address performance limitations on large databases and complex queries~\cite{wang2025macsql}. The framework comprises a core decomposer agent for SQL generation with chain-of-thought reasoning, accompanied by two auxiliary agents that acquire smaller sub-databases and refine erroneous SQL queries. This approach underscores the importance of schema management in text-to-SQL systems, with the selector agent specifically decomposing a large database into a smaller sub-database to mitigate the interference of irrelevant information. BASE-SQL, a pipeline-based method was introduced using open source model fine-tuning, which includes schema linking as its first component, followed by candidate SQL generation, SQL revision, and SQL merge revision~\cite{sheng2025basesql}. This approach represents another architectural paradigm for addressing the schema linking challenge, focusing specifically on cost-effective implementation with open-source models while still achieving competitive performance compared to methods using closed-source models like GPT-4o. In another effort, KaSLA (Knapsack optimization-based schema linking Agent) is proposed, which uses a hierarchical linking strategy and knapsack optimization to select the most relevant schema elements while minimizing redundancy~\cite{yuan2025knapsack}.

Another area of active research in the schema-linking domain is the significant issue with current schema linking evaluation metrics (precision and recall), arguing they fail to adequately capture the impact of missing relevant schema elements. Even a single missing relevant table or column can dramatically decrease SQL generation accuracy while still yielding artificially high performance scores according to standard metrics. The lack of systematic benchmarking for LLM-based text-to-SQL systems is addressed with a comprehensive testing framework with five distinct tasks to evaluate LLM capabilities across the full spectrum of the text-to-SQL process~\cite{zhang2024benchmarking}. An enhanced schema linking metric designed to accurately capture the impact of missing relevant elements, better reflecting actual linking performance and its alignment with SQL generation accuracy, is introduced in~\cite{yuan2025knapsack}. Overall, schema linking represents a critical component in the text-to-SQL pipeline, particularly as the complexity of databases and natural language queries continues to increase. In this work we propose a novel approach to improve the schema-linking and demonstrate how the proposed method shows better results specifically in the customer support domain. We cover this in detail in the Methodology and Experiments sections. As text-to-SQL systems continue to evolve with more sophisticated LLM architectures and multi-agent frameworks, the importance of effective schema linking will likely remain central to achieving robust performance in real-world database environments.

\section{Dataset}
The study utilized two data sources: advertiser-associate conversation data available in the form of chat transcripts and standard operating procedures (SOP) guides used by the support associates to troubleshoot the issues. These SOPs are accessible to all support associates, who undergo comprehensive training to ensure their effective utilization. Additionally, we used advertiser facing self service guides which are rich in educational and troubleshooting steps aimed to be consumed by advertisers directly. Finally, to enhance the Agent's schema linking capabilities, we incorporated domain-specific SQL queries extracted from existing analytics pipelines.

All data underwent thorough pre-processing to remove sensitive information, including personal identifiers, account details and maintaining strict compliance with privacy requirements prior to analysis. While the above datasets originated from Amazon's proprietary databases, their structure follows industry-standard formats similar to customer service data maintained by most companies. Therefore, the methodology described in this paper can be readily adapted to any customer support organization.

\section{Methodology}

\subsection{System Architecture Overview}

The foundation of our approach is a novel multi-agent architecture designed to overcome the fundamental limitations in customer support systems. This specialized division of responsibilities eliminates the traditional bottleneck of manual data source identification, enabling automated schema-linking at enterprise scale. Our multi-agent architecture comprises of three Agents as detailed below:

\begin{itemize}
    \item \textbf{Content Agent:} Extracts structured data requirements from unstructured documentation, including support transcripts, SOPs and advertiser facing help content.
    \item \textbf{Metadata Agent:} Uses enhanced knowledge base that maintains a comprehensive understanding of enterprise database structures, relationships, and usage patterns.
    \item \textbf{Orchestration Agent:} Synthesizes inputs from both agents to generate dynamic schema mapping, SQL queries and coordinates workflow between the above auxiliary Agents.
\end{itemize}

\subsection{Datapoint Discovery}
\begin{figure*}[t]
  \centering
  \includegraphics[width=\textwidth]{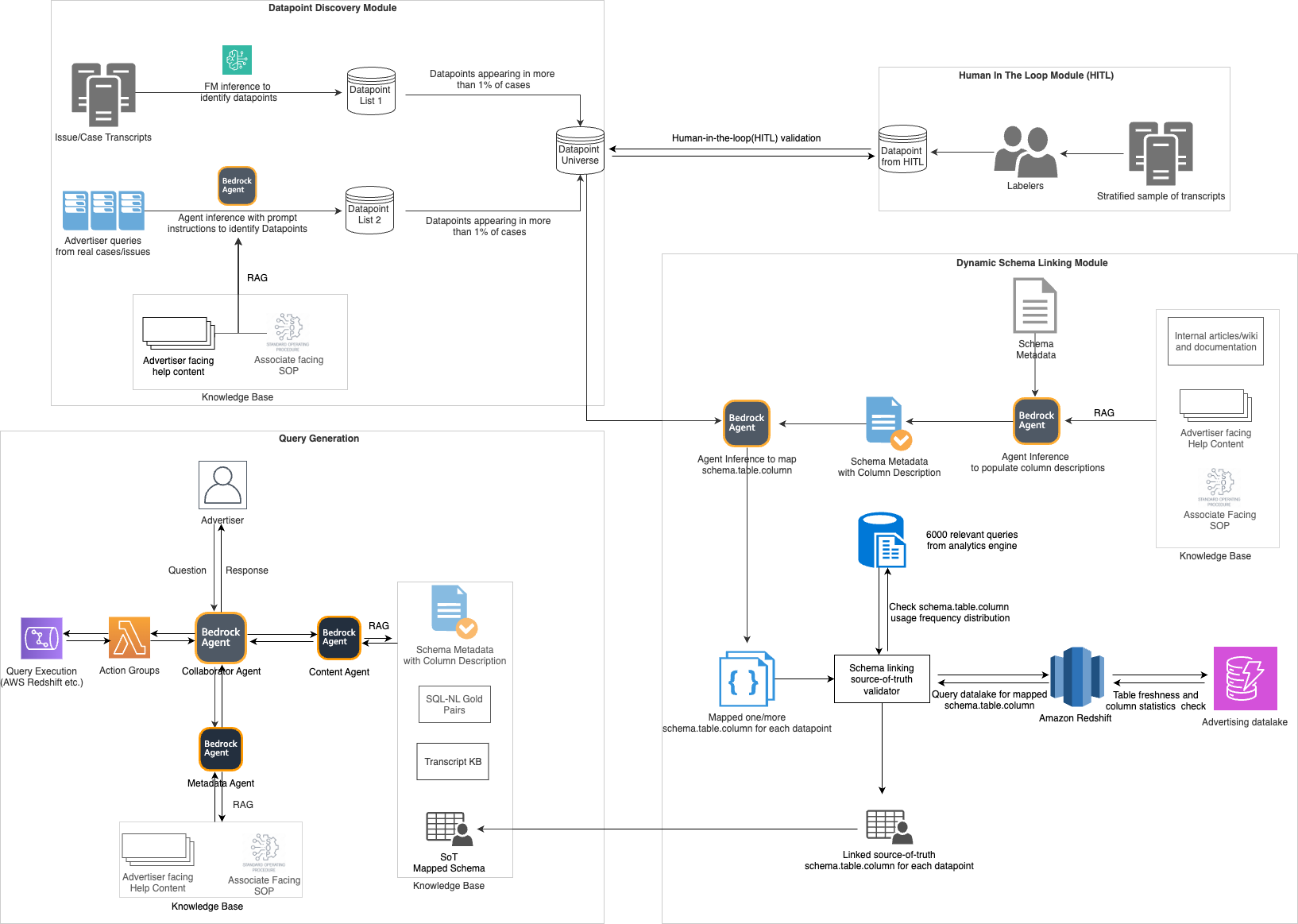}
  \caption{System Architecture and Methodology}
  \label{fig:architecture}
\end{figure*}

As illustrated in Datapoint Discovery module in \cref{fig:architecture}, we followed two pronged approach for datapoint identification. Firstly, we used 20,000 conversations transcripts and prompted an LLM, Claude Sonnet 3.5 V2, to identify datapoints looked up by the associate to resolve the advertiser's question. It is important to note that in production customer service environments, associates often have access to additional platform signals providing customer context, rather than relying solely on direct customer interactions. Accounting for this limitation in transcript data, secondly, we implemented an Agent-based solution incorporating SOPs and advertiser-facing help documentation as the knowledge base and tasked the Agent to answer advertiser's initial question on 5000 stratified sampled cases. Finally, we combined the results of above approaches and retained the common datapoints that occur in at least 1\% of cases. This methodology yielded a comprehensive dictionary of datapoints with their corresponding frequency distributions. The mathematical notation for the approach is captured in the below equations:

\begin{equation}
D_T = \mathcal{F}_{\text{LLM}}(T_{\text{conv}}, P_{\text{id}})
\end{equation}
\begin{equation}
D_A = \mathcal{F}_{\text{Agent}}(Q_{\text{init}}, K_{\text{SOP}}, K_{\text{help}})
\end{equation}
\begin{equation}
f^{\text{norm}}_T(d) = \frac{f_T(d)}{|T_{\text{conv}}|}
\end{equation}
\begin{equation}
f^{\text{norm}}_A(d) = \frac{f_A(d)}{|Q_{\text{init}}|}
\end{equation}
\begin{multline}
D_F = \{d \in (D_T \cup D_A) \mid \\
f^{\text{norm}}_T(d) \geq 0.01 \land f^{\text{norm}}_A(d) \geq 0.01\}
\end{multline}

where; $D_T$, $D_A$ are datapoints from transcript and SOP analysis respectively; $T_{\text{conv}}$ represents 20,000 conversation transcripts; $Q_{\text{init}}$ contains 5000 initial questions; $K_{\text{SOP}}$, $K_{\text{help}}$ are knowledge bases; $P_{\text{id}}$ is the prompt for datapoint identification; $\mathcal{F}_{\text{LLM}}$ and $\mathcal{F}_{\text{Agent}}$ are the functions performed by LLM and Agent respectively; $f_T(d)$ and $f_A(d)$ are raw frequency counts; and $d$ represents an individual datapoint.

\subsection{Dynamic Schema Linking}

\subsubsection{Column Descriptions}
The column names and column data-types alone are not complete to obtain a high accuracy in schema-linking. Specially in enterprise systems, where the different tables are owned by various teams, a standard practice for column names and description is not very common. The ambiguous NL query from the user adds to the complexity. Having a column description for each of the metadata significantly helps in schema-linking. We addressed this challenge by implementing an Agent augmented with a comprehensive knowledge base of internal data lake documentation. The metadata articles, typically available across datalake teams, often exhibit varying degrees of completeness in their documentation. By consolidating these articles with the data point universe discussed in Section 4.2, a comprehensive knowledge repository was established. This enriched context enabled the Agent to generate detailed column descriptions with enhanced accuracy and completeness.

\subsubsection{High Level Schema Mapping}
In this step we used the datapoint universe created in Section 4.2 along with the enhanced column descriptions to generate one-to-many schema mapping for each datapoint. This scalable approach helped us map around 87\% of the datapoints to atleast one schema.table.column in the datalake. One of the existing ambiguity which still prevails in this phase is source of truth determination for scenarios where one datapoint is mapped to multiple schema.table.columns. The below equation describes the high level schema-mapping approach in mathematical terms.

\begin{equation}
S_{\text{map}}(d_i) = \{(s,t,c) \in \mathcal{S} \mid \text{sim}(d_i, C_{\text{desc}}(s,t,c)) > \theta\}
\end{equation}

where; $S_{\text{map}}(d_i)$ is the schema mapping for datapoint $d_i$, $\mathcal{S}$ is the set of all schema.table.column combinations, $\text{sim}()$ is the similarity function, $\theta$ is the similarity threshold.

\subsubsection{Source-of-Truth (SoT) Schema Mapping}
To establish authoritative data sources, we developed a novel multi-factor ranking algorithm that creates a hierarchical decision framework. The system first evaluates data freshness using \texttt{last updated date} fields and table load frequency metrics, then assesses data quality through null value analysis, cardinality evaluation, and finally analyzes usage patterns derived relevant SQL queries. This systematic ranking achieves 84.53\% successful mapping of data points to authoritative data sources, providing a reliable foundation for subsequent SQL query operations. Equation (7) formalizes the SoT schema mapping methodology:

\begin{equation}
\begin{split}
R(s,t,c) &= \alpha F(s,t,c) + \beta Q(s,t,c) + \gamma U(s,t,c) \\
\text{SoT}(d_i) &= \max_{R} S_{\text{map}}(d_i)
\end{split}
\end{equation}

where; $F, Q, U$ represent freshness, quality, and usage scores with weights $\alpha, \beta, \gamma$ respectively ($\alpha + \beta + \gamma = 1$).

\subsection{Query Generation}
SQL query generation accuracy and relevance represent critical success factors in NL-to-SQL systems. The precision of query construction directly impacts the system's ability to retrieve accurate and contextually appropriate data from the underlying database schema. The above steps from data discovery to SoT mapping are crucial in generating the correct SQL. Modern LLMs continue to face challenges with database schema accuracy, often fabricating table and column names even when provided with proper context and instructions. We addressed this limitation by developing a reverse-engineering methodology that converts SQL queries into natural language questions and added it to Agent's knowledge base. The curated SQL-NL pairs were reviewed and validated by analysts. Additionally, we added historical support transcripts into Agent knowledge base to provide real-world conversation examples that can helped to reduce schema hallucination by grounding the Agent's responses. This contextual knowledge helps the Agent to generate more accurate SQL queries by learning from patterns in similar past cases.

\subsection{Human In The Loop (HITL)}
We conducted two distinct HITL validation phases to assess the accuracy of data discovery, schema-linking and SQL generation. In this assessment we leveraged domain experts to analyze a set of stratified sampled cases. The domain experts for these cases are subject matter experts with months of experience in manually resolving the advertiser cases. Additionally, the labelling was parallellized between 10 labellers to reduce human bias. The labelers were tasked to analyze the transcripts along with SOPs to identify the datapoints they needed to resolve the issue manually (without Agentic solution). These ground labels were then used to evaluate the results of data discovery using LLM-as-a-judge. Secondly, we also evaluated the outcomes of schema-linking by tasking analysts to validate SoT and generated SQL query. These analysts have deep familiarity with datalakes and respective tables. Analyst evaluations consisted of binary validation (correct/incorrect) for schema.table.column mappings and Likert scale assessments for generated SQL queries.

\subsection{Evaluation}
For evaluating the system performance we introduce SEAL (Schema Evaluation and Accuracy in Language-to-SQL), a novel composite evaluation metric for NL-to-SQL systems. SEAL incorporates three critical components: datapoint identification accuracy, schema linking precision, and SQL query relevance. The metric is designed to provide a comprehensive assessment while penalizing inconsistent performance across components. The SEAL score is computed as:

\begin{equation}
\begin{split}
\text{SEAL} &= (W_d \cdot A_d^p + W_s \cdot A_s^p + W_q \cdot A_q^p)^{1/p} \\
&\quad \cdot \left(1 - \frac{\sigma}{\mu \cdot k}\right) \\
&\text{where } W_d + W_s + W_q = 1
\end{split}
\end{equation}

where; $A_d, A_s, A_q$ are accuracy scores (datapoints, schema, SQL); $p < 1$ is penalty parameter; $\sigma, \mu$ are std.dev and mean; $k = 3$.

Unlike traditional harmonic mean, SEAL incorporates a penalty parameter $p$ that explicitly penalizes performance variance across components, providing finer control over score aggregation. In this implementation, the weights were configured as $W_d = 0.3$, $W_s = 0.4$, and $W_q = 0.3$, allocating slightly higher importance to schema-linking accuracy while maintaining overall balance. The penalty parameter was set to $p = 0.5$ to ensure robust penalization of under performing components.

\section{Experiments}
The implementation utilized a multi-agent architecture comprising one collaborative agent and two auxiliary agents: the Content Agent and Metadata Agent. Five distinct experimental configurations were designed, focusing on knowledge base enrichment and prompt enhancement to optimize schema-linking performance. \cref{tab:results} presents a comparative analysis of these experiments against the established baseline. The experimental design maintains the Content Agent as a control while systematically evaluating various feature enhancements of the Metadata Agent. Performance evaluation employed four key metrics: datapoint identification accuracy, schema-linking accuracy, SQL relevance score, and SEAL score. In the baseline evaluation, we passed the NL query in the multi-agent setup to get a SEAL score of 76.57. In enhancement 1, we included the datapoint universe in the prompt instructions with objective to restrict fabrication/hallucination in identified datapoints. We observed that this improves the datapoint accuracy by 0.81\% and SEAL score by 1.99\%. In enhancement 2, we further added the SoT schema.table.column for each datapoint (derived from dynamic schema linking) to the metadata agent KB. Enhancement 2 led to the most uplift in SEAL score improving it by 3.66\% and schema-linking accuracy by 2.79\% against baseline. In enhancement 3, we added the transcripts to the Metadata KB resulting in improvements of 7.99\% in SQL accuracy and 5.25\% in SEAL score respectively. The best performance was achieved after including the reverse engineered SQL-NL pairs in the knowledge base. Our proposed framework, enhancement-4, achieves a 6.65\% improvement in SEAL score compared to the baseline implementation, demonstrating significant performance gain.

\begin{table*}[t]
  \caption{Experimental Results Across Different Feature Configurations}
  \label{tab:results}
  \centering
  \begin{small}
  \begin{tabular}{lcccc}
    \toprule
    \textbf{Experiment} & \textbf{Data-Point} & \textbf{Schema} & \textbf{SQL} & \textbf{SEAL} \\
    & \textbf{Accuracy} & \textbf{Accuracy} & \textbf{Accuracy} & \\
    \midrule
    Baseline (F1) & 75.02 & 79.84 & 78.35 & 76.57 \\
    Enhancement 1 (F1+F2) & 75.63 & 80.02 & 79.05 & 78.09 \\
    Enhancement 2 (F1+F2+F3) & 78.13 & 82.07 & 80.76 & 79.37 \\
    Enhancement 3 (F1+F2+F3+F4) & 79.53 & 84.13 & \textbf{84.61} & 80.59 \\
    Enhancement 4 (F1+F2+F3+F4+F5) & \textbf{80.96} & \textbf{84.53} & 82.09 & \textbf{81.66} \\
    \bottomrule
  \end{tabular}
  \end{small}
  \vskip 0.5ex
  \footnotesize
  \begin{flushleft}
    F1: Metadata with Column Description; F2: Datapoint Discovery; F3: Dynamic Schema Linking; F4: Transcript KB; F5: SQL-NL Pairs
  \end{flushleft}
\end{table*}

\begin{figure*}[t]
  \centering
  \includegraphics[width=\textwidth]{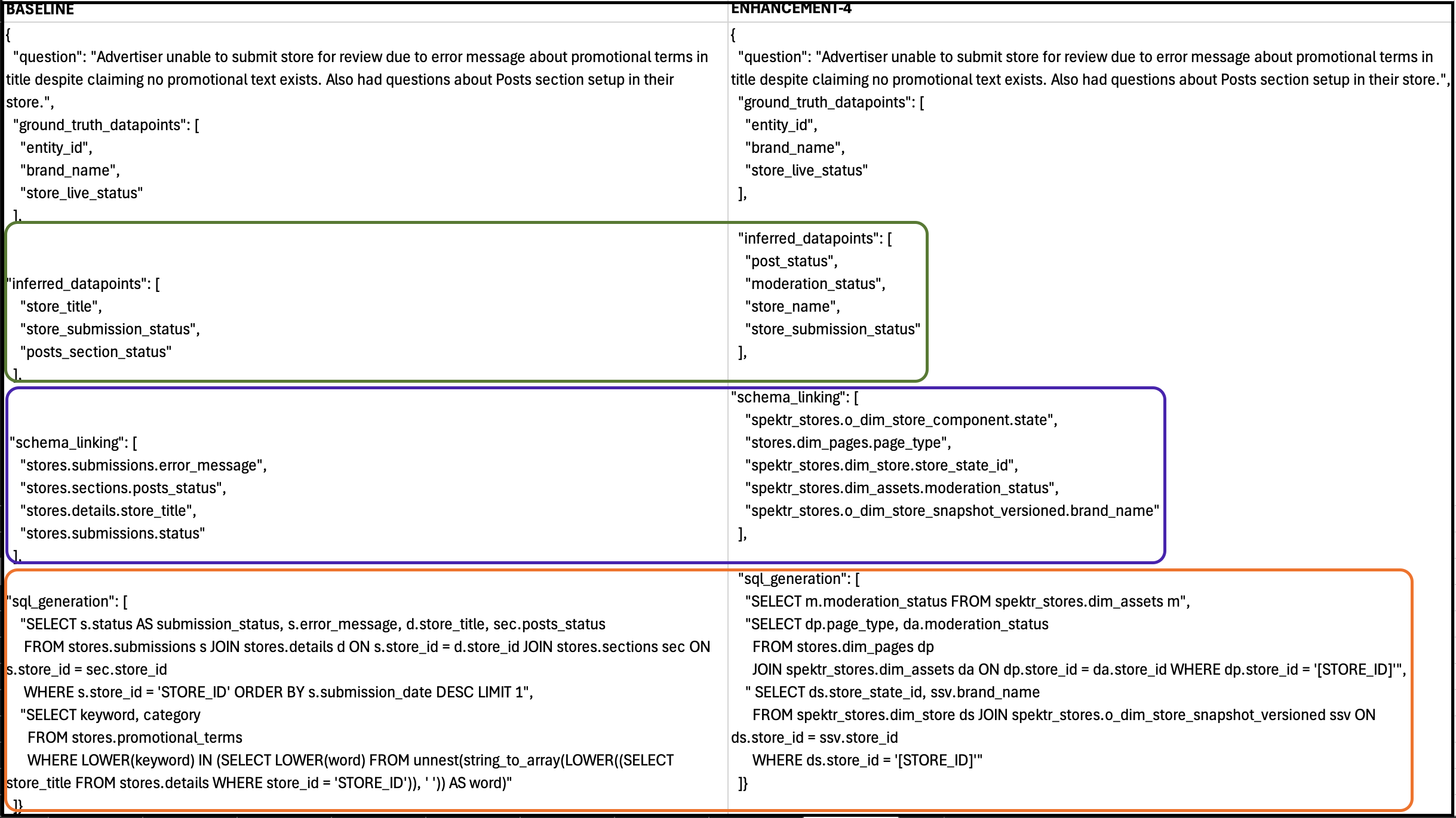}
  \caption{Baseline vs. Proposed Approach: Real-World Example Results}
  \label{fig:comparison}
\end{figure*}

As illustrated in \cref{fig:comparison}, the experimental results demonstrate a significant improvement in query generation accuracy and schema identification when comparing baseline and enhanced feature sets through color-coded comparisons. In the baseline, we use the schema metadata with column descriptions as the only knowledge base to observe the performance and experiment with further enhancements as highlighted in \cref{tab:results}. Analysis of the baseline implementation reveals critical limitations, particularly in schema comprehension, where the agent exhibited severe hallucinations. In \cref{fig:comparison}, the green box shows comparison between data points identified by the baseline vs the enhancement 4. The baseline agent hallucinates \texttt{posts\_\allowbreak section\_\allowbreak status} based on a mere mention of ``Posts section'' in the advertiser query. On the contrary, in the proposed approach, the data point is rightly identified as \texttt{post\_\allowbreak status}. The blue box shows comparison of data points getting mapped to respective schema.table.column. The baseline agent generates invalid mappings with non-existent tables and columns, which leads to errors in SQL generation. On other hand, Agent with enhancement-4 successfully identifies correct schema paths by inferring necessary dependent columns (e.g., \texttt{spektr\_\allowbreak stores.\allowbreak dim\_\allowbreak store.\allowbreak store\_\allowbreak state\_\allowbreak id}, \texttt{spektr\_\allowbreak stores.\allowbreak dim\_\allowbreak pages.\allowbreak page\_\allowbreak type}).

Finally, the orange box showcases improvements in SQL generation. The baseline agent attempts to reference imaginary tables such as \texttt{stores.\allowbreak promotional\_\allowbreak terms}, while enhancement 4 demonstrates marked improvements in query precision and schema recognition, accurately identifying specific dimensional tables (e.g.,\texttt{spektr\_\allowbreak stores.\allowbreak dim\_\allowbreak store\_\allowbreak component}) instead of using generic \texttt{stores} references. Notable improvements also include proper identification of ``moderation status`` through \texttt{spektr\_\allowbreak stores.\allowbreak dim\_\allowbreak assets.\allowbreak moderation\_\allowbreak status} and accurate mapping of post-related information using \texttt{store\_\allowbreak state\_\allowbreak id} and \texttt{page\_\allowbreak type} attributes. The Agent with enhancement~4 correctly utilizes the dimensional model structure, replacing hallucinated relationships with precise join conditions and actual table references. We observed similar improvements and patterns consistently across multiple examples, reinforcing the robustness and generalizability of our approach.

\section{Conclusion and Future Work}
Our experiments demonstrates that SAFAARI successfully addresses critical challenges in schema linking for enterprise-scale advertising support systems, achieving significant improvements in accuracy and development efficiency. The achieved results include 80.96\% accuracy in datapoint identification, 84.53\% accuracy in schema-linking, and an 8× improvement in development efficiency which validate the framework's effectiveness in real-world applications. Our approach achieves 6.65\% improvement in SEAL score, a novel metric introduced in this work for NL-to-SQL systems, against the baseline. Future work can focus on these key areas: (1) validating the framework's performance across diverse customer service domains to establish broader applicability benchmarks, (2) extending the SEAL metric to incorporate dynamic performance indicators such as query execution efficiency and computational resource optimization (3) investigating the integration of reinforcement learning techniques to improve schema linking accuracy through continuous learning from user interactions.

\bibliography{sources}
\bibliographystyle{icml2026}

\end{document}